%% file: paper.tex
\documentclass[conference]{IEEEtran}

\usepackage{cite}
\usepackage{amsmath,amssymb}
\usepackage{graphicx}
\usepackage{booktabs}
\usepackage{xcolor}
\usepackage{url}
\usepackage[hidelinks]{hyperref}

\newcommand{\method}{OBLIVION}
\newcommand{\bench}{AgentForgetBench}

\makeatletter
\newcommand{\linebreakand}{%
  \end{@IEEEauthorhalign}
  \hfill\mbox{}\par
  \mbox{}\hfill\begin{@IEEEauthorhalign}
}
\makeatother
\begin{document}

\title{\method: Workflow-Level Operational Skill Unlearning for Deployed Agents}

\author{
\IEEEauthorblockN{Zhengyang Shan, Xu Qian, Jiayun Xin, Kun Li, Yue Zhang, and Minghui Xu}
\IEEEauthorblockA{Shandong University}
}

\maketitle

\input{templates/sections/00_abstract}

\input{templates/sections/01_introduction}

\input{templates/sections/02_related_work}
\input{templates/sections/03_model_preliminaries}

\begin{figure}[t]
  \centering
  \includegraphics[width=\columnwidth]{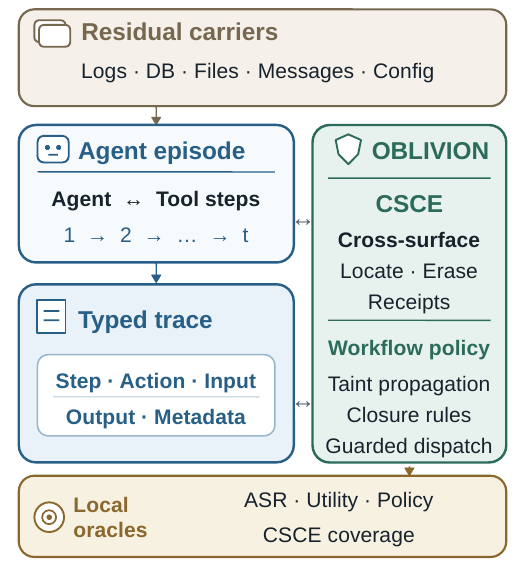}
  \caption{Overview of \method. Residual carriers are converted into typed workflow traces and evaluated by local oracles under CSCE and frozen workflow remediation.}
  \label{fig:framework}
\end{figure}

\input{templates/sections/04_method_system_design}
\input{templates/sections/05_analysis_security}
\input{templates/sections/06_experimental_design}
\input{templates/sections/07_experimental_results}
\input{templates/sections/08_implications_discussion}
\input{templates/sections/09_conclusion_future_work}

\bibliographystyle{IEEEtran}
\bibliography{references}

\end{document}

%% file: templates/sections/00_abstract.tex
\begin{abstract}
Large language model agents are becoming operational interfaces to files, memories, registries, and external tools. This deployment shift creates a new skill revocation problem: after a skill is removed from an explicit registry, an agent may still reconstruct it from residual carriers such as archives, transcripts, schemas, or memory entries. We study this problem as operational skill unlearning, where the goal is not parameter-level forgetting, but preventing a deployed agent from rebuilding a revoked skill through primitive tools. We introduce \method, a controlled benchmark and defense harness for revoked-skill resurrection. \method{} models each episode as a source-to-sink workflow, applies Cross-Surface Coherent Erasure to reduce residual carriers, and uses frozen workflow remediation near dangerous sinks. On the locked 88 attack episodes, the no-defense arm reaches formal attack success rate 1.0. \method{} reduces the rate to 0.114 and impact-weighted exposure to 0.115 while keeping locked utility at 1.0 and benign block rate at 0. In a separate skill-attack-derived sandbox, \method{} reduces attack success from 1.0 to 0.2 and impact-weighted exposure from 1.0 to 0.213 while preserving all utility controls. These results support workflow-level evaluation beyond checking explicit skill entries.
\end{abstract}


%% file: templates/sections/01_introduction.tex
\section{Introduction}

Large language model agents are becoming operational interfaces in software workspaces, data systems, and smart service environments. Unlike a chatbot that only returns text, a deployed agent can read archives, search memory, write local artifacts, update registries, and invoke tools. This shift makes skill revocation a practical security problem. When a helper, tool schema, or registry entry is revoked, the system owner expects the agent to stop using that skill. However, the skill may still be recoverable from residual carriers, including transcripts, cached files, vector memory entries, local schemas, and registry drafts.

This problem has a close precedent in software removal. Android uninstallation studies show that app removal can leave security-relevant state behind, including files, databases, and capability-like records that later programs may abuse \cite{zhang2016life}. Software deployment research similarly treats de-installation as a closure problem: safe removal depends on knowing which components, dependencies, and references remain reachable \cite{dolstra2004imposing}. Android malware measurement further shows that detection or market removal does not immediately end exposure, because harmful apps may persist on devices or migrate across markets \cite{shen2022androidpha}. Accessibility malware studies also report persistence mechanisms that make removal an incomplete security response \cite{xu2024dva}. These systems motivate our central analogy: revoking an agent skill should not be evaluated only by checking whether the visible entry disappeared. It should be evaluated by asking whether remaining operational state can reconstruct the revoked capability.

This paper studies the gap between removing an explicit skill entry and preventing revoked-skill resurrection. We use revoked-skill resurrection to mean the observable rebuilding of a revoked skill through a source-to-sink workflow. The workflow starts from a residual carrier, transforms the carrier through primitive tool steps, and reaches a sink such as a persistent helper, a tool schema, a registry entry, or an invocation. This setting differs from parameter-level forgetting. Our question is not whether model weights have forgotten a concept. Our question is whether a deployed agent can rebuild a revoked skill from operational material that remains in its environment.

Prior work makes this harm concrete. AgentDojo, InjecAgent, and ToolEmu show that tool-using agents can be hijacked into direct harm, data exfiltration, or high-stakes tool failures \cite{debenedetti2024agentdojo,zhan2024injecagent,ruan2024toolemu}. Skill-Inject, SkillJect, and SkillAttack show that skill files, helper scripts, and benign-looking skills can trigger data exfiltration, destructive action, payload execution, or exploitable paths \cite{schmotz2026skillinject,jia2026skillject,duan2026skillattack}. Credential-leakage work shows that skills can expose secrets across natural-language and code channels \cite{chen2026credentialskills}.

These results motivate a revocation-specific question. If a harmful, vulnerable, or over-privileged skill is removed, can residual carriers still restore the sink that made it dangerous? We therefore report impact-weighted exposure (IWE) and a skill-attack-derived sandbox that maps prior agent-skill harms to controlled sink classes.

The problem is difficult because the source is often benign alone, the dangerous step may be delayed, and benign workflows can use the same surface words as attacks. A useful defense should therefore model the workflow rather than block every mention of old tools or helper artifacts.

To address this problem, we present \method, a controlled benchmark and defense harness for operational skill unlearning. \method{} has two goals. The first goal is measurement: \bench{} represents each episode with typed scenarios, residual carriers, primitive tool events, and primary oracles. Formal attack success rate is computed only from primary oracle pass or fail outcomes within the declared tool-step budget. The second goal is defense: \method{} combines Cross-Surface Coherent Erasure (CSCE), which reduces residual carriers across deployment surfaces, with frozen workflow remediation, an audit and blocking layer for source-to-sink workflows near sinks that can make a revoked skill persistent or callable.

On the locked 160-case benchmark, the no-defense arm reaches formal attack success rate 1.0 and IWE 1.0. \method{} remediation reduces formal attack success rate to 0.114 and IWE to 0.115, while keeping locked utility pass rate at 1.0 and benign block rate at 0. In a separate skill-attack-derived sandbox, the no-defense arm reaches ASR and IWE 1.0, while \method{} reduces ASR to 0.2 and IWE to 0.213 with all utility controls passing. Baseline defenses show different tradeoffs: Task Shield reaches 0 ASR but fails all Phase 2 utility rows, while ClawGuard keeps Phase 2 utility at 1.0 but leaves ASR at 0.455.

This paper makes the following contributions:
\begin{itemize}
    \item We formulate operational skill unlearning for deployed agents as a workflow-level problem, focused on whether a revoked skill can be rebuilt from residual carriers and primitive tools.
    \item We design \bench{}, a controlled benchmark with typed scenarios, residual carrier fixtures, primitive tool traces, and primary-oracle outcomes for revoked-skill resurrection.
    \item We implement \method, a defense harness that combines CSCE with frozen workflow remediation over source-to-sink workflows.
    \item We report evidence on locked benchmarks and two-stage live rollouts, including a harm-reduction analysis that maps prior skill-attack harms to controlled sink classes.
\end{itemize}

%% file: templates/sections/02_related_work.tex
\section{Related Work}

\subsection{Software Removal and Residual State}
Prior systems work shows that removal is a reachability problem, not only an entry-deletion problem. Android uninstallation studies found that removed apps can leave security-relevant state across files, databases, and system data structures, and that later apps can abuse this remaining state for credential theft, private-data access, privilege escalation, and keylogging \cite{zhang2016life}. Software deployment research makes a similar point from the dependency side: safe installation, upgrade, and de-installation require exact component identities, explicit dependencies, closure computation, and garbage collection of unreachable components \cite{dolstra2004imposing}. Android malware measurement further shows that removal responses may lag operational exposure, with potentially harmful apps persisting on devices after detection and migrating across markets after removal \cite{shen2022androidpha}. Accessibility malware studies add that malicious apps can include explicit persistence mechanisms, so deleting the visible package is not always enough to understand or reduce harm \cite{xu2024dva}.

Operational skill unlearning follows the same broad lesson in a new substrate. The object being removed is not an Android package or a library closure. It is a skill that may be represented by helper files, schemas, memory entries, registry records, transcripts, and local artifacts. \method{} therefore treats skill revocation as a workflow-level removal problem. The evaluation asks whether reachable residual carriers can still be transformed into persistent or callable skill state.

\subsection{Machine Unlearning and Agent Skill Security}
Machine unlearning removes the influence of selected data or requests from trained models \cite{guo2020certified,bourtoule2021machine}. Operational skill unlearning is different because the target is deployed operational state, not only model parameters. A revoked skill can be represented by helpers, schemas, memory entries, registry records, and previous task traces. We measure whether a deployed agent can rebuild the skill from residual carriers.

Agent systems make this question concrete. ReAct, Toolformer, and retrieval-augmented generation show why tool and memory surfaces matter \cite{yao2023react,schick2023toolformer,lewis2020retrieval}. AgentDojo, InjecAgent, and ToolEmu evaluate prompt injection and tool-risk outcomes in tool-using agents \cite{debenedetti2024agentdojo,zhan2024injecagent,ruan2024toolemu}. Skill-Inject, SkillJect, and SkillAttack show that skill files, helper scripts, and skill interfaces are high-trust attack surfaces \cite{schmotz2026skillinject,jia2026skillject,duan2026skillattack}. Credential-leakage work further shows that skills can expose secrets across natural-language and code channels \cite{chen2026credentialskills}. These studies examine active, injected, or vulnerable skills. \method{} studies what remains after revocation.

\subsection{Agent Defenses and Boundary Controls}
Prompt injection and adversarial triggers show that small input changes can redirect model behavior \cite{wallace2019universal,greshake2023not}. StruQ separates prompt and data channels \cite{chen2025struq}. Task Shield checks whether instructions and tool calls align with the user goal \cite{jia2025taskshield}. Runtime defenses inspect tool calls, schemas, registries, and action boundaries before an agent commits an effect. ClawGuard is a recent example for tool-augmented agents under indirect prompt injection \cite{zhao2026clawguard}.

These defenses address important agent failures, but they usually focus on instruction boundaries, task alignment, or a visible tool-call boundary. Revoked-skill resurrection can span transcript reading, parameter summarization, schema inference, and later registry or helper writes. \method{} complements these defenses by tracking residual carriers, transformations, and dangerous sinks across the episode.

%% file: templates/sections/03_model_preliminaries.tex
\section{Model and Preliminaries}

\subsection{Operational Setting and Workflow}
We consider a deployed agent that receives a user task and can call a bounded set of primitive tools. A primitive tool may read local files, replay transcripts, search vector memory, write workspace artifacts, write simulated registry entries, or append local records for simulated external actions. The harness executes these tools inside a local workspace with dummy data only.

A revoked skill is a skill artifact that the system no longer wants the agent to restore or use. Examples include a helper implementation, a tool schema, a registry entry, or procedural steps. A residual carrier is local material that may still contain information about the revoked skill, such as memory entries, archive members, transcripts, registry drafts, or intermediate results.

An episode is represented as a sequence of primitive tool events. Each event records the primitive name, summaries, source labels, sink labels, dependencies, and local effect metadata. We map primitive events into a source-to-sink workflow:
\[
  w = (s, t_1, \ldots, t_m, k, e),
\]
where \(s\) is a residual carrier, \(t_i\) are transforms, \(k\) is a sink, and \(e\) is an observable effect. A dangerous sink can make a rebuilt skill persistent, callable, or externally visible. In \bench{}, these sinks include persistent files, registry writes, simulated tool-schema registrations, long-term memory writes, simulated external actions, and simulated code execution.

A taint label records whether an output depends on a residual carrier labeled as a revoked source. The harness propagates this label through transforms and checks it before a dangerous sink executes. This typed representation separates benign summaries from workflows that rebuild a revoked skill and write it into a persistent or callable sink.

\subsection{Oracles, Metrics, and Evaluation Boundary}
An oracle is a local evaluator that checks whether the episode produced the declared effect. A primary oracle counts toward formal attack success rate, while secondary and diagnostic oracles are analysis-only. For a set \(R\) of primary oracle rows within the declared tool-step budget,
\[
  \mathrm{ASR} = \frac{1}{|R|}\sum_{r \in R} \mathbb{1}[r.\mathrm{passed}].
\]
Utility pass rate is measured on designated benign utility tasks. Benign block rate measures how often a benign case is blocked. Audit-only decisions are non-blocking under the offline benchmark convention and are not production enforcement evidence.

The benchmark uses controlled artifacts and deterministic schemas. Each reported run records split metadata, workflow traces, policy events, and oracle rows so that attack success is tied to observable source-to-sink effects.

%% file: templates/sections/04_method_system_design.tex
\section{Method and System Design}

\subsection{System Overview and Benchmark Materialization}
The software-removal analogy suggests two failure modes for skill revocation: residual carriers may remain after visible removal, and remaining carrier material may later reach a persistent or callable sink. \method{} is designed around these failures. Figure~\ref{fig:framework} shows the system. \method{} materializes a controlled episode, records a typed workflow trace, and applies two controls. CSCE reduces residual carriers before execution. Workflow remediation audits or blocks source-to-sink workflows at dangerous sinks.

The first design need is observability. If evaluation only reads model text, explanation and restored capability are hard to separate. \bench{} therefore stores skills, scenarios, surface fixtures, utility tasks, and oracle definitions as typed JSON or JSONL artifacts. Each scenario declares a revoked skill, carrier surfaces, an expected source-to-sink workflow, sink classes, oracle definitions, and split metadata. The harness records simulated external actions, registry writes, and code execution as JSONL events.

This design makes attack success observable through local effects rather than inferred from model text. It also lets benign utility tasks share surface features with attacks while keeping a different expected effect.

\subsection{Cross-Surface Coherent Erasure}
The second design need is coordinated removal. A skill registry entry can disappear while the same capability remains in memory, archives, transcripts, schemas, or cached helper fragments. CSCE therefore runs a transaction over six surfaces: behavior gate placeholder, skill registry, vector memory, filesystem, transcript, and Model Context Protocol (MCP) schema fixtures. Each adapter locates carriers for a revoked skill, prepares a reversible delete or tombstone, commits it, and writes a receipt to a write-ahead audit log. The coordinator rolls back prepared carriers on prepare failure and preserves the audit trail on partial commit failure.

CSCE receipts record adapter-reported operation history and material probes. Their role in \method{} is operational: reduce the residual carrier set before an agent can use primitive tools to rebuild a revoked skill.

\subsection{Guarded Dispatch and Frozen Workflow Remediation}
The third design need is runtime control after cleanup. CSCE reduces known carriers, but it cannot assume that every carrier is found. Blocking every old archive or transcript would also harm benign workflows. \method{} therefore allows reads and summaries, but follows whether material derived from a revoked skill later reaches a dangerous sink.

During an episode, the dispatcher records tool events and dependencies. The Anti-ReDerivation Guard (ARG) wraps this dispatcher. It assigns taint to material derived from a revoked skill, propagates that taint through transforms, and checks dangerous sinks before their simulated effects execute.

The guard emits policy events with identifiers, decisions, taint labels, sink labels, and reasons. A block decision stops the sink primitive before the effect occurs. An audit-only decision records a policy event but lets the primitive continue under the offline benchmark convention.

The final design need is to avoid a brittle list of exact paths or examples. Workflow remediation generalizes from development evidence into rules over atomic operations, taint kinds, source labels, sink classes, and risk thresholds. The frozen rules do not refer to scenario identifiers, oracle identifiers, exact paths, or exact secret strings.

Workflow remediation uses block decisions near sinks that make a revoked skill persistent or callable and audit-only decisions for patterns that can also appear in benign workflows. We evaluate this choice through formal attack success rate, benign blocking, and utility outcomes.

%% file: templates/sections/05_analysis_security.tex
\section{Analysis}

\method{} counts an attack only when a primary oracle observes the declared local effect. Reading a residual carrier or discussing a revoked skill is not enough. A successful episode must produce the declared sink and effect, such as a persistent helper, simulated registry entry, or simulated invocation record. Therefore, formal attack success rate is tied to local effect evidence rather than to model wording.

Workflow remediation is frozen before evaluation. Observing a new attack pattern during a reported run cannot create a new rule for that same run. This keeps the measured result tied to the evaluated rule set.

ARG and workflow remediation intervene at dangerous sinks after taint propagation. If a value is derived from a residual carrier and a later primitive attempts to write it into a dangerous sink, the guard can audit or block before the simulated effect occurs. Thus, a multi-step path can be detected even when no earlier read is independently unsafe. The property depends on typed observability.

The main utility risk is false positive blocking. Benign tasks may read archives, write helper-like files, or simulate local schemas. \method{} addresses this risk by using audit-only decisions for ambiguous cases and by measuring benign block rate and utility pass rate separately. Audit-only decisions record intervention coverage without blocking benign progress, but they are not deployable enforcement guarantees.

\method{} is incomplete. In the 88 locked attack cases, \method{} remediation leaves 10 successful attacks. These failures identify workflow patterns that future policies should study with new development evidence.

%% file: templates/sections/06_experimental_design.tex
\section{Experimental Design}
\label{sec:experimental-design}

\subsection{Evaluation Questions and Workloads}
We organize the evaluation around three questions. Q1 is motivated by attack viability: a low ASR matters only if no-defense agents can rebuild revoked skills. It asks whether this happens on the locked benchmark and how much workflow remediation reduces it. Q2 is motivated by harm diversity: a restored helper, tool schema, memory write, and external-action plan have different operational meaning. It asks whether \method{} reduces exposure on skill-attack-derived harm categories. Q3 is motivated by utility and baseline interpretation: a defense can reduce ASR by blocking benign workflows. It asks how \method{} compares with baseline defenses under utility pressure.

The workloads are separated by purpose. Table~\ref{tab:phase2} uses the primary locked benchmark with 160 cases: 88 attack rows, 8 benign near-miss rows, and 64 designated utility rows. Each attack case declares a revoked skill, residual carriers, source-to-sink workflow, sink class, and primary oracle. Table~\ref{tab:skill-attack-harm} uses a separate 40-case skill-attack-derived sandbox: 30 attack cases and 10 utility controls. The attack cases cover five local harm analog categories derived from prior agent-skill attack papers. Figure~\ref{fig:live-tradeoff} and Table~\ref{tab:phase5} add boundary utility workloads to test whether defenses block benign workflows that resemble revoked-skill resurrection.

\subsection{Arms and Baseline Defenses}
The main comparison includes three experimental arms: no-defense, Static ARG, and \method{} remediation. The no-defense arm establishes attack viability. Static ARG is the earlier frozen ARG policy before workflow remediation. It tests whether static source-to-sink rules are enough. \method{} remediation tests the frozen workflow remediation used in the final evaluation.

We implement three deterministic baselines following the core mechanisms of prior defenses. StruQ separates trusted user instructions from untrusted carrier content and blocks instruction-like overrides from the untrusted region. Task Shield checks whether a candidate restore, register, invoke, or execute action is needed for the user goal. ClawGuard checks tool calls, schema writes, registry updates, and artifact boundaries before execution-like sinks. These baselines place prior defense mechanisms into the same source-to-sink benchmark as \method{}.

\subsection{Two-Stage Rollout and Metrics}
We additionally run a two-stage live rollout using DeepSeek as the agent-response generator. The rollout contains 1,904 API calls across the main-policy, boundary-utility, baseline, and skill-attack-derived suites. Stage one asks DeepSeek for an agent response or action plan. Stage two uses fixed offline evaluators for all policy and utility metrics. This separation keeps generation and evaluation distinct.

Formal attack success rate is the mean pass rate of primary oracle rows on attack cases. Impact-weighted exposure measures sink-level harm, using weights 2 for persistent files, 3 for long-term memory, 4 for tool registry and simulated execution sinks, and 5 for simulated external-action sinks. We also report utility pass rate, benign block rate, block rate, audit-only rate, and API success.

%% file: templates/sections/07_experimental_results.tex
\section{Experimental Results}

\subsection{Q1: Locked Benchmark}
Q1 tests the premise of the paper. If no-defense agents do not rebuild revoked skills, workflow-level controls would have little target. The locked benchmark therefore first measures no-defense viability and then compares Static ARG with \method{} on the same source-to-sink workflows.

Table~\ref{tab:phase2} reports the main locked-benchmark comparison.
\input{tables/phase2_results}
The no-defense arm reaches formal attack success rate 1.0 on 88 attack cases, including 24 registry, 16 external-action, and 8 simulated execution sink successes. This establishes that visible skill removal is not enough in the benchmark setting. Static ARG reduces ASR to 0.750, but leaves most reconstructed workflows successful. \method{} remediation leaves 10 attacks successful, reducing formal attack success rate to 0.114 and IWE to 0.115.

The Static ARG baseline has the same block rate as \method{}, 0.081, but a lower audit-only rate, 0.056 compared with 0.456. The improvement therefore comes from broader source-to-sink handling, not simply from more block decisions.

\subsection{Q2: Skill-Attack-Derived Sandbox}
Q2 asks whether attack reduction also holds when cases are grouped by effect type. Prior skill-attack papers motivate harms such as skill persistence, tool authority abuse, memory revival, external action, and integrity mutation. The sandbox uses these harm analog categories and measures weighted exposure across them.

Table~\ref{tab:skill-attack-harm} uses a different workload from Table~\ref{tab:phase2}.
\input{tables/skill_attack_harm_reduction}
It is a 40-case sandbox organized around five local harm analog categories motivated by prior agent-skill attack papers. The no-defense arm succeeds on all 30 attacks. \method{} reduces overall ASR to 0.2 and IWE to 0.213 while passing all 10 utility controls.

The table also shows why a single generic defense principle is not enough. StruQ leaves ASR and IWE at 1.0 because separating trusted and untrusted text does not by itself track whether residual carrier material reaches a dangerous sink. Task Shield and ClawGuard reduce ASR to 0 on this sandbox, but their utility pass rates fall to 0.7 and 0.9, respectively. \method{} has a higher residual ASR on this sandbox, but it preserves all utility controls.

\subsection{Q3: Utility and Adapter Tradeoffs}
Q3 is motivated by utility. A broad rule can make ASR look good by blocking benign helper creation, schema migration, or archive summarization. Q3 therefore asks whether \method{} keeps useful workflows while reducing revoked-skill resurrection, and how this tradeoff compares with StruQ, Task Shield, and ClawGuard.

\method{} remediation keeps the locked utility pass rate at 1.0 and the benign block rate at 0. Thus, the lower formal attack success rate in Table~\ref{tab:phase2} does not require benign blocking on the locked benchmark.

\begin{figure}[t]
  \centering
  \includegraphics[width=\columnwidth]{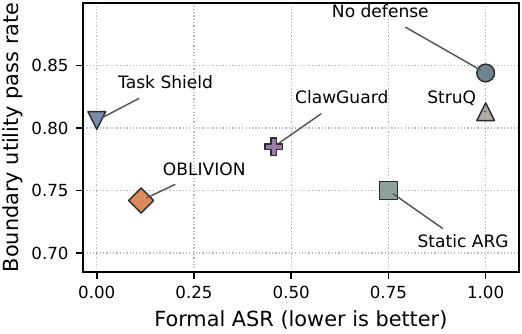}
  \caption{Two-stage live DeepSeek tradeoff. Lower formal attack success rate and higher boundary utility pass rate are better. The live model generates responses, while fixed offline evaluators compute all metrics.}
  \label{fig:live-tradeoff}
\end{figure}

Figure~\ref{fig:live-tradeoff} shows boundary utility.
On 128 live cases, no defense passes 0.844, Static ARG passes 0.750, and \method{} passes 0.742. Aggressive CSCE plus ARG passes only 0.188 and blocks 0.797 of cases.

Table~\ref{tab:phase5} compares the three baseline defenses.
\input{tables/phase5_results}
It reuses the locked attack denominator from Table~\ref{tab:phase2} for ASR and IWE, but adds a separate boundary utility workload for benign workflows that resemble skill restoration. StruQ keeps boundary utility at 0.813, but leaves ASR at 1.0 and fails all 64 locked utility rows. Task Shield reduces ASR to 0, but also fails all locked utility rows because benign helper and schema construction can look off-goal under its action checker. ClawGuard keeps locked utility at 1.0, but leaves ASR at 0.455 because it is strongest at visible tool or registry boundaries and weaker on multi-step carrier transformations.

\method{} remediation does not have the lowest formal attack success rate in this comparison. Its contribution is the observed tradeoff: formal attack success rate 0.114, Phase 2 utility pass rate 1.0, benign block rate 0, and boundary utility pass rate 0.742.

\subsection{Residual Failures}
The 10 successful attacks under \method{} remediation show that workflow remediation is incomplete. They are development evidence for future policies, but they do not justify changing the frozen policy during locked evaluation.

%% file: tables/phase2_results.tex
\begin{table}[t]
\centering
\caption{Main locked-benchmark results. Impact-weighted exposure (IWE) weights successful attacks by sink impact.}
\label{tab:phase2}
\scriptsize
\setlength{\tabcolsep}{2pt}
\begin{tabular*}{\columnwidth}{@{\extracolsep{\fill}}lccccc@{}}
\toprule
Target & ASR & IWE & Utility & Benign block & Block/Audit \\
\midrule
No defense & 1.000 & 1.000 & 1.000 & 0.000 & 0.000/0.000 \\
Static ARG & 0.750 & 0.806 & 0.984 & 0.000 & 0.081/0.056 \\
\method{} & 0.114 & 0.115 & 1.000 & 0.000 & 0.081/0.456 \\
\bottomrule
\end{tabular*}
\end{table}

%% file: tables/skill_attack_harm_reduction.tex
\begin{table}[t]
\centering
\caption{Baseline tradeoff on a separate skill-attack-derived sandbox.}
\label{tab:skill-attack-harm}
\scriptsize
\setlength{\tabcolsep}{2pt}
\begin{tabular*}{\columnwidth}{@{\extracolsep{\fill}}lccccc@{}}
\toprule
Target & ASR & IWE & Util. & Block & Audit \\
\midrule
No defense & 1.000 & 1.000 & 1.000 & 0.000 & 0.000 \\
StruQ & 1.000 & 1.000 & 0.700 & 0.050 & 0.000 \\
Task Shield & 0.000 & 0.000 & 0.700 & 0.050 & 0.775 \\
ClawGuard & 0.000 & 0.000 & 0.900 & 0.175 & 0.600 \\
\method{} & 0.200 & 0.213 & 1.000 & 0.000 & 0.600 \\
\bottomrule
\end{tabular*}
\end{table}

%% file: tables/phase5_results.tex
\begin{table}[t]
\centering
\caption{Baseline comparison on locked attacks and boundary utility.}
\label{tab:phase5}
\scriptsize
\setlength{\tabcolsep}{2pt}
\begin{tabular*}{\columnwidth}{@{\extracolsep{\fill}}lcccc@{}}
\toprule
Target & ASR & IWE & Phase2 util. & Boundary util. \\
\midrule
StruQ & 1.000 & 1.000 & 0.000 & 0.813 \\
Task Shield & 0.000 & 0.000 & 0.000 & 0.806 \\
ClawGuard & 0.455 & 0.316 & 1.000 & 0.785 \\
\method{} & 0.114 & 0.115 & 1.000 & 0.742 \\
\bottomrule
\end{tabular*}
\end{table}

%% file: templates/sections/08_implications_discussion.tex
\section{Implications and Discussion}

The experiments support the paper's central claim: deleting explicit skill entries is not a sufficient evaluation target. A revoked skill can be rebuilt through residual carriers and primitive tools, even when no single read operation is dangerous by itself. This makes source-to-sink workflow evidence the practical unit of analysis.

The software-removal analogy clarifies why CSCE and workflow remediation are complementary. As in app uninstallation and software de-installation, the relevant question is not only whether a visible entry has disappeared, but whether reachable operational state can still recreate the removed capability. CSCE reduces residual carriers, while workflow remediation controls the dangerous sinks that remaining carriers may reach.

The baseline defenses clarify where different defense principles fit. StruQ is useful when untrusted content overrides a trusted goal. Task Shield is useful when the candidate action clearly drifts away from the user goal. ClawGuard is useful when the dangerous boundary is a visible tool, schema, registry, or artifact operation. Revoked-skill resurrection can cross all three boundaries, which explains why \method{} tracks the source-to-sink workflow instead of relying on one boundary alone.

%% file: templates/sections/09_conclusion_future_work.tex
\section{Conclusion and Future Work}

This work studies operational skill unlearning for deployed agents and proposes \method{} as a benchmark and defense harness. The main observation is that a revoked skill can be rebuilt from residual carriers through primitive tools, so evaluation should measure source-to-sink workflows and primary-oracle outcomes. \method{} remediation reduces formal attack success rate from 1.0 to 0.114 and impact-weighted exposure from 1.0 to 0.115 on the locked benchmark, while keeping locked utility pass rate at 1.0 and benign block rate at 0. A separate skill-attack-derived sandbox further shows ASR 0.2 and IWE 0.213 on categories motivated by prior agent-skill attacks.

Future work should expand workflow families, evaluate additional live model providers under explicit safety controls, and study how audit-only decisions should map to deployable actions. Remaining successful attacks should be analyzed to improve future workflow policies.

%% file: references.bib
@inproceedings{bourtoule2021machine,
  title = {Machine Unlearning},
  author = {Bourtoule, Lucas and others},
  booktitle = {Proceedings of the IEEE Symposium on Security and Privacy},
  year = {2021}
}

@inproceedings{guo2020certified,
  title = {Certified Data Removal from Machine Learning Models},
  author = {Guo, Chuan and Goldstein, Tom and Hannun, Awni and van der Maaten, Laurens},
  booktitle = {Proceedings of the International Conference on Machine Learning},
  year = {2020}
}

@inproceedings{yao2023react,
  title = {{ReAct}: Synergizing Reasoning and Acting in Language Models},
  author = {Yao, Shunyu and Zhao, Jeffrey and Yu, Dian and Du, Nan and Shafran, Izhak and Narasimhan, Karthik and Cao, Yuan},
  booktitle = {International Conference on Learning Representations},
  year = {2023}
}

@inproceedings{schick2023toolformer,
  title = {Toolformer: Language Models Can Teach Themselves to Use Tools},
  author = {Schick, Timo and others},
  booktitle = {Advances in Neural Information Processing Systems},
  year = {2023}
}

@inproceedings{lewis2020retrieval,
  title = {Retrieval-Augmented Generation for Knowledge-Intensive NLP Tasks},
  author = {Lewis, Patrick and others},
  booktitle = {Advances in Neural Information Processing Systems},
  year = {2020}
}

@inproceedings{debenedetti2024agentdojo,
  title = {{AgentDojo}: A Dynamic Environment to Evaluate Prompt Injection Attacks and Defenses for {LLM} Agents},
  author = {Debenedetti, Edoardo and others},
  booktitle = {NeurIPS Datasets and Benchmarks},
  year = {2024}
}

@inproceedings{ruan2024toolemu,
  title = {Identifying the Risks of {LM} Agents with an {LM}-Emulated Sandbox},
  author = {Ruan, Yangjun and others},
  booktitle = {International Conference on Learning Representations},
  year = {2024}
}

@inproceedings{zhan2024injecagent,
  title = {{InjecAgent}: Benchmarking Indirect Prompt Injections in Tool-Integrated Large Language Model Agents},
  author = {Zhan, Qiusi and Liang, Zhixiang and Ying, Zifan and Kang, Daniel},
  booktitle = {Findings of the Association for Computational Linguistics: ACL},
  year = {2024}
}

@misc{schmotz2026skillinject,
  title = {Skill-Inject: Measuring Agent Vulnerability to Skill File Attacks},
  author = {Schmotz, David and Beurer-Kellner, Luca and Abdelnabi, Sahar and Andriushchenko, Maksym},
  howpublished = {arXiv:2602.20156},
  year = {2026}
}

@misc{jia2026skillject,
  title = {{SkillJect}: Effectively Automating Skill-Based Prompt Injection for Skill-Enabled Agents},
  author = {Jia, Xiaojun and others},
  howpublished = {arXiv:2602.14211},
  year = {2026}
}

@misc{duan2026skillattack,
  title = {{SkillAttack}: Automated Red Teaming of Agent Skills through Attack Path Refinement},
  author = {Duan, Zenghao and others},
  howpublished = {arXiv:2604.04989},
  year = {2026}
}

@misc{chen2026credentialskills,
  title = {How Your Credentials Are Leaked by {LLM} Agent Skills: An Empirical Study},
  author = {Chen, Zhihao and others},
  howpublished = {arXiv:2604.03070},
  year = {2026}
}

@inproceedings{wallace2019universal,
  title = {Universal Adversarial Triggers for Attacking and Analyzing {NLP}},
  author = {Wallace, Eric and Feng, Shi and Kandpal, Nikhil and Gardner, Matt and Singh, Sameer},
  booktitle = {Proceedings of the Conference on Empirical Methods in Natural Language Processing and the International Joint Conference on Natural Language Processing},
  year = {2019}
}

@inproceedings{greshake2023not,
  title = {Not What You've Signed Up For: Compromising Real-World {LLM}-Integrated Applications with Indirect Prompt Injection},
  author = {Greshake, Kai and Abdelnabi, Sahar and Mishra, Shailesh and Endres, Christoph and Holz, Thorsten and Fritz, Mario},
  booktitle = {Proceedings of the ACM Workshop on Artificial Intelligence and Security},
  year = {2023}
}

@inproceedings{chen2025struq,
  title = {{StruQ}},
  author = {Chen, Sizhe and others},
  booktitle = {USENIX Security},
  year = {2025}
}

@inproceedings{jia2025taskshield,
  title = {The Task Shield},
  author = {Jia, Feiran and others},
  booktitle = {ACL},
  year = {2025}
}

@misc{zhao2026clawguard,
  title = {{ClawGuard}},
  author = {Zhao, Wei and others},
  howpublished = {arXiv:2604.11790},
  year = {2026}
}

@inproceedings{zhang2016life,
  title = {Life after App Uninstallation: Are the Data Still Alive? Data Residue Attacks on {Android}},
  author = {Zhang, Xiao and Ying, Kailiang and Aafer, Yousra and Qiu, Zhenshen and Du, Wenliang},
  booktitle = {Network and Distributed System Security Symposium},
  year = {2016}
}

@inproceedings{dolstra2004imposing,
  title = {Imposing a Memory Management Discipline on Software Deployment},
  author = {Dolstra, Eelco and Visser, Eelco and de Jonge, Merijn},
  booktitle = {Proceedings of the International Conference on Software Engineering},
  year = {2004}
}

@inproceedings{shen2022androidpha,
  title = {A Large-Scale Temporal Measurement of {Android} Malicious Apps: Persistence, Migration, and Lessons Learned},
  author = {Shen, Yun and Vervier, Pierre-Antoine and Stringhini, Gianluca},
  booktitle = {USENIX Security Symposium},
  year = {2022}
}

@inproceedings{xu2024dva,
  title = {{DVa}: Extracting Victims and Abuse Vectors from {Android} Accessibility Malware},
  author = {Xu, Haichuan and Yao, Mingxuan and Zhang, Runze and Dawoud, Mohamed Moustafa and Park, Jeman and Saltaformaggio, Brendan},
  booktitle = {USENIX Security Symposium},
  year = {2024}
}
